\documentclass[conference]{IEEEtran}
\IEEEoverridecommandlockouts
\usepackage{cite}
\usepackage{amsmath,amssymb,amsfonts}
\usepackage{algorithmic}
\usepackage{graphicx}
\usepackage{textcomp}
\usepackage{xcolor}
\usepackage{multirow}
\usepackage{array}
\def\BibTeX{{\rm B\kern-.05em{\sc i\kern-.025em b}\kern-.08em
    T\kern-.1667em\lower.7ex\hbox{E}\kern-.125emX}}

\begin{document}

\title{WiFlow: A Lightweight WiFi-based Continuous Human Pose Estimation Network with Spatio-Temporal Feature Decoupling}

\author{
\IEEEauthorblockN{Yi Dao\IEEEauthorrefmark{1}, 
    Lankai Zhang\IEEEauthorrefmark{1}, 
    Hao Liu\IEEEauthorrefmark{1}, 
    Haiwei Zhang\IEEEauthorrefmark{2} and
    Wenbo Wang\IEEEauthorrefmark{1}
    }
    \IEEEauthorblockA{\IEEEauthorrefmark{1}Faculty of Mechanical and Electrical Engineering, Kunming University of Science and Technology}
    \IEEEauthorblockA{\IEEEauthorrefmark{2}Shanghai ChingMu Vision Technology Co. Ltd}
    \IEEEauthorblockA{  
         \mbox{\{daoy1, lankzhang, liuhao666 \}@stu.kust.edu.cn},
        \mbox{zhanghaiwei@chingmu.com},
        \mbox{wenbo\_wang@kust.edu.cn}
    }
    \vspace*{-8mm}
}

\maketitle

\begin{abstract}
Human pose estimation is fundamental to intelligent perception in the Internet of Things (IoT), enabling applications ranging from smart healthcare to human-computer interaction. While WiFi-based methods have gained traction, they often struggle with continuous motion and high computational overhead. This work presents WiFlow, a novel framework for continuous human pose estimation using WiFi signals. Unlike vision-based approaches such as two-dimensional deep residual networks that treat Channel State Information (CSI) as images, WiFlow employs an encoder-decoder architecture. The encoder captures spatio-temporal features of CSI using temporal and asymmetric convolutions, preserving the original sequential structure of signals. It then refines keypoint features of human bodies to be tracked and capture their structural dependencies via axial attention. The decoder subsequently maps the encoded high-dimensional features into keypoint coordinates. Trained on a self-collected dataset of 360,000 synchronized CSI-pose samples from 5 subjects performing continuous sequences of 8 daily activities, WiFlow achieves a Percentage of Correct Keypoints (PCK) of 97.25\% at a threshold of 20\% (PCK@20) and 99.48\% at PCK@50, with a mean per-joint position error of 0.007 m. With only 2.23M parameters, WiFlow significantly reduces model complexity and computational cost, establishing a new performance baseline for practical WiFi-based human pose estimation. Our code and datasets are available at https://github.com/DY2434/WiFlow-WiFi-Pose-Estimation-with-Spatio-Temporal-Decoupling.git.
\end{abstract}


\section{Introduction}
Human Pose Estimation (HPE) requires accurately perceiving and quantifying the structure, joint positions, and overall morphology of the human body in three-dimensional space. By identifying the spatial coordinates of human keypoints, this technology constructs a digital representation of the human skeleton, demonstrating significant application value in numerous fields such as elderly care~\cite{xu2023robust} and immersive Virtual Reality (VR)~\cite{kotaru2017position}. 
Currently, mainstream HPE research primarily relies on visual or wearable devices. Vision-based models such as OpenPose~\cite{cao2019openpose} have achieved high-precision estimation but are constrained by lighting conditions and privacy concerns. Wearable solutions~\cite{huang2020deepfuse} avoid privacy issues but require intrusive devices. In contrast, WiFi sensing utilizing Channel State Information (CSI) offers low cost, non-contact operation, and privacy preservation. As WiFi sensing has achieved breakthrough progress in tasks such as human activity recognition~\cite{yi2024probsparse}\cite{luo2024vision}, respiratory monitoring\cite{fan2024contactless}\cite{alzaabi2025design}, and gesture recognition\cite{yan2025wi}~\cite{jeong2024ubigest}, researchers have begun to explore its application in the more challenging HPE task.

WiFi-based HPE is essentially a complex regression problem that nonlinearly maps CSI signal variations to human keypoint coordinates. To solve this, researchers have explored various architectural paradigms. The pioneering work WiSPPN~\cite{wang2019can} investigated the feasibility of using WiFi for replacing visual tasks by regressing a Pose Adjacent Matrix (PAM) through deep residual networks, which encode both absolute keypoint positions and their relative spatial constraints. Following this, approaches like Wi-Mose~\cite{wang2021point} and WPNet~\cite{yang2022metafi} adopted a direct regression strategy, using deep convolutional networks to predict keypoint coordinates directly from CSI features. While this simplified the inference pipeline, it came at the cost of potentially losing structural constraint information inherent in the human skeleton.

To address the limitations of simple CNNs, recent research has introduced more sophisticated designs. PerUnet~\cite{zhou2022perunet} combines U-Net's multi-scale feature fusion capability with Transformer's global modeling ability to regress PAM. By dynamically focusing on subcarriers sensitive to posture changes through multi-head attention mechanisms, it significantly improves estimation accuracy. Similarly, MetaFi++~\cite{zhou2023metafi++} adopts a multi-branch architecture to process spatial features of different antenna pairs through shared CNNs, reducing parameter complexity while preserving spatial diversity. Its subsequent Transformer encoder adaptively captures key feature patterns across subcarriers.

However, despite these advancements, existing methods still face fundamental challenges. First, they often treat CSI data simply as images, ignoring its inherent temporal characteristics and physical significance. The temporal dimension of CSI has strict causal constraints, while the subcarrier dimension characterizes the spatial distribution of frequency responses. 2D CNNs may conflate these dimensions, leading to loss of key information. For instance, WiPose~\cite{jiang2020towards} attempts to extract spatial and temporal features separately through a CNN-LSTM cascade architecture, but the CNN preprocessing may disrupt the integrity of the original temporal structure. CSI-former~\cite{zhou2022csi} adopts a pure Transformer encoder that achieves global modeling but has shortcomings in capturing local spatial correlations and suffers from high computational complexity.

More critically, existing research is mostly based on evaluating discrete pose samples, lacking systematic research on continuous action sequences. Unlike image data, the information density of WiFi CSI frames is relatively sparse, and single frame signals often cannot fully characterize pose features, requiring feature aggregation within time windows. In practical applications, human pose changes are often continuous and smooth motion processes. Modeling this temporal continuity is crucial for improving system practicality and reducing the "jitter" often seen in frame-by-frame predictions. Additionally, current methods lack deep consideration of CSI sensing characteristics when designing confidence weighting strategies; simply borrowing confidence from the visual domain may introduce incompatible prior biases.

To address this, we propose WiFlow, a deep learning framework specifically optimized for continuous human pose estimation. The core innovation of WiFlow lies in explicitly modeling the spatio-temporal coupled features of CSI signals: capturing the temporal dynamics of CSI through TCN~\cite{bai2018empirical}, and extracting spatial correlations between subcarriers using asymmetric convolution while maintaining temporal structure integrity. Compared to traditional deep residual networks, this architecture achieves more efficient feature extraction while significantly reducing model parameters (Params) and floating-point operations (FLOPs). Furthermore, we employ an axial attention mechanism~\cite{wang2020axial} that achieves joint modeling of intra-keypoint feature aggregation and inter-keypoint dependencies. 
The main contributions of this paper are as follows:
\begin{enumerate}
    \item We construct and release a continuous human pose WiFi sensing dataset containing 360,000 precisely synchronized CSI-pose sample pairs, covering continuous motion sequences of multiple angles and action types.
    \item We propose the WiFlow model, which achieves explicit decoupling and efficient extraction of CSI spatio-temporal features through a TCN-asymmetric CNN collaborative encoding architecture, and realizes feature screening within keypoints and dependency modeling between keypoints through axial attention mechanisms.
    \item We systematically explore network architecture design principles specifically tailored for continuous WiFi CSI-based human pose estimation. Experiments show that WiFlow achieves 97.25\% accuracy on PCK@20 with MPJPE of only 0.007 m, using only 2.23 million parameters—significant improvements over baseline methods in both accuracy and efficiency.
\end{enumerate}

\section{WiFlow Architecture}

\subsection{CSI Data Collection and Preprocessing}
In Orthogonal Frequency Division Multiplexing (OFDM) systems, CSI characterizes the propagation of wireless signals from the transmitter to the receiver. 
Mathematically, the Channel Frequency Response (CFR) for the $k$-th subcarrier at time $t$ can be modeled as:
\begin{equation}
H(k, t) = \|H(k, t)\| e^{j \angle H(k, t)},
\end{equation}
where $\|H(k, t)\|$ and $\angle H(k, t)$ represent the amplitude and phase response, respectively. 
The received signal is a superposition of multipath components, which can be decomposed into a static component $H_s$ (background) and a dynamic component $H_d$ (human motion):
\begin{equation}
H(k, t) = H_s(k) + \sum_{n \in \mathcal{P}_{dynamic}} \alpha_n(t) e^{-j \frac{2\pi d_n(t)}{\lambda}},
\end{equation}
where $H_s(k)$ represents the static reflections (e.g., Line-of-Sight and walls), and $\mathcal{P}_{dynamic}$ denotes the set of dynamic paths reflected by human body parts. 
For the $n$-th dynamic path, $\alpha_n(t)$ is the complex attenuation factor, $d_n(t)$ is the propagation path length, and $\lambda$ is the carrier wavelength.
Human movement induces changes in $d_n(t)$, causing fluctuations in CSI, which serves as the physical basis for WiFi sensing.

In this work, we utilize the Intel 5300 Network Interface Card (NIC) with the Linux CSI Tool~\cite{halperin2011tool}. 
The system operates at a center frequency of 5 GHz with a 20 MHz bandwidth. 
To fully capture CSI signals reflected from different angles, we configure 3 transmitting antennas and 2 receivers (each with 3 receiving antennas). 
This setup generates $3 \times 3 = 9$ communication links per receiver, totaling $9 \times 2 = 18$ links across the system. 
The CSI sampling frequency is 600 Hz, synchronized with 30 FPS video data (used for supervision), and we adopt a sliding window of sample size $T=20$.

For preprocessing, we utilize only the amplitude information $\|H(k, t)\|$ and discard the phase. 
While phase contains distance information, it is heavily corrupted by Carrier Frequency Offset (CFO) and Sampling Frequency Offset (SFO) in commercial WiFi devices. 
Sanitizing raw phase data requires computationally expensive algorithms, which contradicts our goal of designing a lightweight, edge-deployable framework. 
Furthermore, empirical studies~\cite{miao2025wi} suggest that amplitude variations induced by body scattering are sufficiently discriminative for pose estimation in rich-scattering indoor environments.
We integrate data from all 18 links along the subcarrier dimension. Since each link contains 30 subcarriers, the final network input tensor is $\mathbf{X} \in \mathbb{R}^{540 \times 20}$.

\subsection{Human Pose Estimation Based on Spatio-Temporal Feature Extraction}

\begin{figure}[t]
\centerline{\includegraphics[width=\columnwidth]{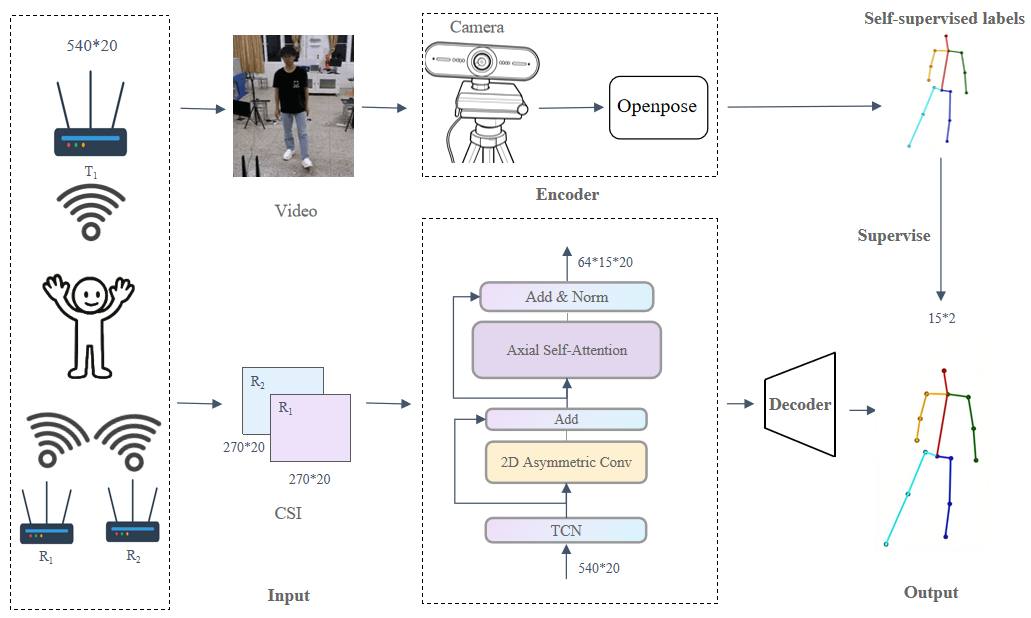}}
\caption{WiFlow network architecture diagram}
\label{architecture}
\end{figure}

WiFlow employs an encoder-decoder architecture for continuous WiFi-based pose estimation. We leverage OpenPose to extract keypoint annotations from synchronized videos as automatic training labels. Unlike traditional autoencoders that compress and reconstruct data within the same modality, our encoder-decoder framework performs supervised cross-modal translation, namely, converting wireless signal patterns into geometric keypoint representations, which can be generated by vision tools such as OpenPose. Considering this is a regression task and encoded features already possess explicit semantics, we employ convolutional decoding for efficient coordinate regression. As shown in Fig.~\ref{architecture}, the encoder extracts features in three stages: the first stage uses dilated causal convolution in the TCN module to extract temporal features and screen subcarriers; the second stage employs asymmetric residual blocks to extract spatial features, compressing the subcarrier dimension to the keypoint number; the third stage introduces axial attention to reinforce key features along the width direction and model inter-keypoint dependencies along the height direction. Finally, the decoder maps encoded features to keypoint coordinates.

\subsubsection{\textbf{Temporal Convolutional Network for Temporal Feature Extraction}}

CSI data is essentially a multivariate time series where temporal and subcarrier dimensions exhibit fundamental asymmetry: the temporal dimension has strict causal constraints reflecting dynamic changes over time, while the subcarrier dimension characterizes the spatial distribution of frequency responses. 2D CNN may conflate these dimensions, disrupting temporal structure integrity. Therefore, we adopt a spatiotemporal decoupling strategy, extracting temporal features before spatial modeling. 

For temporal feature extraction, the key is maintaining causality while capturing global dependencies. Common temporal modeling methods include LSTM, Transformer, and TCN. While LSTM can handle long-range dependencies, its sequential processing nature results in computational inefficiency and susceptibility to gradient vanishing. Transformer offers high parallelism but imposes high computational complexity of $O(L^2)$,  which is often excessive for the short, fixed-length sequences typical in CSI windows. In contrast, TCN, through causal and dilated convolutions, enables parallel computation with controllable receptive fields. Given these considerations, we adopt TCN as the temporal encoder. TCN combines 1D fully convolutional networks with causal convolutions, maintaining consistent output length through zero padding. The core dilated convolution operation is defined as:
\begin{equation}
F(s) = (\mathbf{x} *_d w)(s) = \sum_{i=0}^{k-1} w(i) \cdot \mathbf{x}_{s-d \cdot i},
\end{equation}
where $\mathbf{x}$ is the input sequence, $w$ is the convolution kernel, $s$ is the temporal index of the output sequence, $d$ is the dilation factor, $k$ is the kernel size, $w(i)$ represents the kernel weight at position $i$, and $s-d \cdot i$ ensures only information at or before time $s$ is accessed, thus preserving causality. By setting the dilation rate $d$ to grow exponentially across layers, the network can cover the entire sequence with fewer layers.

Additionally, while multi-receiver systems provide multi-angle reflection information of human joints, not all subcarriers provide effective pose-related information. Therefore, we employ a progressive channel compression strategy, simultaneously extracting temporal features and implementing subcarrier screening. Specifically, our network utilizes a decoupled architecture combining grouped convolutions and $1\times 1$ pointwise convolutions. The progressive compression and screening are mathematically formulated as a cross-channel projection operated by the pointwise convolution:
\begin{equation}
X^{(l+1)} = \sigma(\mathbf{W}_{1\times 1}^{(l)} * \mathcal{F}_{group}(X^{(l)}) + \mathbf{b}^{(l)}),
\end{equation}
where $\mathcal{F}_{group}(\cdot)$ represents the temporal feature extraction within $G$ independent subcarrier groups (implemented via the dilated causal convolutions defined in Equation 3), and $\mathbf{W}_{1\times 1}^{(l)}$ is the $1\times 1$ pointwise convolution kernel. This projection is subject to a monotonically decreasing channel constraint $C^{(l+1)} < C^{(l)}$ across $L$ temporal blocks. Through end-to-end optimization, the pointwise kernel $\mathbf{W}_{1\times 1}^{(l)}$ acts as a learnable cross-group screening operator. It adaptively assigns smaller weights to noise-dominant frequency components, thereby effectively filtering out subcarriers weakly correlated with human poses while fusing and retaining significant temporal patterns in a more compact feature subspace.

\subsubsection{\textbf{Asymmetric Convolutional Network for Spatial Feature Extraction}}

After TCN processing, the next step requires extracting spatial correlations between subcarriers. The core challenge in spatial feature extraction is mapping high-dimensional subcarrier features to semantically explicit keypoint representations. Standard 2D convolution kernels operate simultaneously on temporal and spatial dimensions, disrupting the temporal structure preserved by TCN and introducing unnecessary computational overhead. Therefore, we design an asymmetric convolutional network using $1 \times k$ kernels to focus exclusively on spatial feature extraction in the subcarrier dimension while keeping the temporal dimension unchanged. Following the same design principle of U-Net, the network expands channel numbers while progressively compressing the subcarrier dimension to preserve information capacity. For input features $\mathbf{X} \in \mathbb{R}^{B \times C_n \times T \times S}$, where $B$ is batch size, $C_n$ is the input channel number of the $n$-th residual layer, $T$ is CSI frame length, and $S$ is the number of subcarriers, the transformation of the $n$-th residual block is defined as:
\begin{equation}
\mathbf{X}^{n+1} = \sigma\left(\sum_{i=1}^{3}\mathbf{W}_i^n * \mathbf{X}_i^n + \mathbf{b}_i^n\right) + \mathbf{W}_s^n * \mathbf{X}^n,
\end{equation}
where $\mathbf{W}_i^n \in \mathbb{R}^{C_{n+1} \times C_n \times 1 \times k}$ is the asymmetric convolution kernel of layer $i$ operating only in the subcarrier dimension, $*$ denotes the convolution operation, $\sigma(\cdot)$ is the SiLU activation function, and $\mathbf{W}_s^n \in \mathbb{R}^{C_{n+1} \times C_n \times 1 \times 1}$ is the downsampling convolution with stride $(1,2)$. This progressive mapping achieves hierarchical transformation from $S$ subcarriers to $K$ semantic keypoints, with each output dimension corresponding to a human keypoint.

\subsubsection{\textbf{Axial Self-Attention}}

After spatiotemporal feature encoding, the model obtains feature representation $\mathbf{X} \in \mathbb{R}^{B \times C \times K \times T}$, where $K$ corresponds to the keypoint dimension. At this point, each feature position has established correspondence with human keypoints, but explicit structured dependency modeling between $K$ keypoints is still lacking, and features of each keypoint need further screening for final coordinate regression. Therefore, we introduce the Axial Self-Attention mechanism, achieving refined feature processing through attention computation in two orthogonal directions. Unlike Vision Transformer's approach of flattening 2D features into 1D sequences, axial attention significantly reduces computational complexity through a decomposition strategy while maintaining spatial topology. Specifically, axial attention decomposes 2D self-attention into two independent 1D attention operations along height and width axes, reducing computational complexity from $O(H^2W^2)$ to $O(H^2W + HW^2)$.

Our axial attention adopts a two-stage processing strategy: the first stage computes attention along the width direction, modeling dependencies between internal features of each keypoint to highlight important features. We reshape the input to $(B \times K) \times C \times T$ form, allowing features of $B \times K$ keypoints to be processed in parallel, and generate query, key, and value matrices through learnable linear transformations:
\begin{equation}
\mathbf{Q} = \mathbf{XW}_Q, \mathbf{K} = \mathbf{XW}_K, \mathbf{V} = \mathbf{XW}_V,
\end{equation}
where $\mathbf{W}_Q, \mathbf{W}_K, \mathbf{W}_V \in \mathbb{R}^{T \times T}$ are weight matrices, and $\mathbf{Q}, \mathbf{K}, \mathbf{V} \in \mathbb{R}^{(B \times K) \times C \times T}$. To enhance the model's ability to capture diverse feature patterns, axial attention adopts a grouping mechanism. We divide the $T$-dimensional features into $G$ independent groups with $d=T/G$ features per group, allowing different groups to focus on different types of feature dependencies. Within each group, attention between internal keypoint features is calculated through scaled dot product:
\begin{equation}
\text{Attention}^{\text{width}}(\mathbf{Q}, \mathbf{K}, \mathbf{V}) = \text{softmax}\left(\frac{\mathbf{Q}^T\mathbf{K}}{\sqrt{d}}\right)\mathbf{V}^T,
\end{equation}
where $\mathbf{Q}, \mathbf{K}, \mathbf{V} \in \mathbb{R}^{d \times T}$. After concatenating the outputs of $G$ groups, the first-stage output is obtained through a batch normalization layer and reshaped back to the original dimension of $B \times C \times K \times T$. Unlike standard attention computation, we calculate the correlation between positions $i$ and $j$ through dot product of feature values across all channels. This cross-channel aggregation not only computes correlations between internal keypoint features but also derives this correlation by comprehensively considering features from all channels.

The second stage computes attention along the height direction, modeling structured dependencies between $K$ keypoints. The attention computation at this stage is similar to the first stage, except the input is reshaped to $(B \times T) \times C \times K$. Through this approach, the model effectively captures feature correlations within keypoints and structured dependencies between keypoints, providing rich global context information for subsequent pose decoding.

\subsubsection{\textbf{Decoder Layer}}

The decoder progressively maps high-dimensional features to final 2D keypoint coordinates through multi-layer convolution. It consists of two stages: the feature refinement stage uses $3 \times 3$ convolution for dimension reduction with batch normalization and SiLU activation to enhance nonlinear expression, then $1 \times 1$ convolution compresses channels to 2, corresponding to keypoint x and y coordinates; the coordinate regression stage compresses the feature map from $(2, K, T)$ to $(2, K, 1)$ through adaptive average pooling. Compared to traditional fully connected layer regression, pooling aggregation avoids numerous parameters while achieving lightweight design. Finally, squeeze and transpose operations produce the $K \times 2$ output tensor, corresponding to 2D coordinates of $K$ keypoints.

\subsubsection{\textbf{Evaluation Metrics and Loss Function}}

In the experiments, we selected the Percentage of Correct Keypoints (PCK), widely used in human pose recognition tasks, as the evaluation metric, defined as:
\begin{equation}
\text{PCK}@\alpha = \frac{1}{N \times K}\sum_{i=1}^{N}\sum_{j=1}^{K} I\left(\frac{\|\hat{p}_{ij} - p_{ij}\|_2}{s_i} \leq \alpha\right),
\end{equation}
where $\hat{p}_{ij}$ and $p_{ij}$ represent the predicted and ground truth coordinates of the $j$-th keypoint of the $i$-th sample respectively, $N$ is the number of samples, $K$ is the number of keypoints, $I(\cdot)$ is an indicator function that returns 1 when the condition is satisfied and 0 otherwise, $s_i$ is the normalization factor. We use the distance from right shoulder to left hip as the reference scale, and the threshold $\alpha$ determines the strictness of evaluation. Smaller $\alpha$ values evaluate the model's high-precision prediction capability, while larger $\alpha$ values evaluate the model's overall coverage. We use multiple thresholds $\alpha \in \{0.1, 0.2, 0.3, 0.4, 0.5\}$ for comprehensive evaluation in experiments. Additionally, we use Mean Per Joint Position Error (MPJPE) as an auxiliary metric:
\begin{equation}
\text{MPJPE} = \frac{1}{N \times K}\sum_{i=1}^{N}\sum_{j=1}^{K}\|\hat{p}_{ij} - p_{ij}\|_2.
\end{equation}
This metric directly reflects the average Euclidean distance between predicted and ground truth coordinates, which is important for evaluating the visual coherence of pose sequences.

For the loss function, we adopt the direct coordinate regression strategy. Although PAM can provide structured constraints, experiments show that direct regression with appropriate loss design also achieves excellent performance. Notably, although the confidence information provided by visual models like OpenPose reflects detection reliability and focusing on high-confidence keypoints can prevent learning unreliable labels, we found that directly using confidence weighting does not always improve CSI pose estimation performance, as visually difficult keypoints (such as occluded parts) may be clearly visible to WiFi signals, and vice versa. Therefore, we adopt a loss function without confidence weighting, allowing the model to autonomously learn the CSI-to-pose mapping. We define the training process as minimizing the error between predicted and ground truth keypoints. Considering outlier impacts, we use Smooth L1-norm loss as the main loss function:
\begin{equation}
\mathcal{L}_H = \frac{1}{N}\sum_{i=1}^{N}\frac{1}{K}\sum_{j=1}^{K}\|\hat{p}_{ij} - p_{ij}\|_H,
\end{equation}
where $\|\cdot\|_H$ denotes the Smooth L1 norm. For vector $x = (x_1, x_2)$, it is defined as:
\begin{equation}
\|x\|_H = h(x_1) + h(x_2),
\end{equation}
where the Smooth L1-norm regularization term is defined as:
\begin{equation}
h(x) = \begin{cases}
\frac{0.5x^2}{\beta} & \text{if } |x| < \beta, \\
|x| - 0.5\beta & \text{if } |x| \geq \beta,
\end{cases}
\end{equation}
where $\beta$ is the smoothing parameter of Smooth L1-norm loss. This loss provides smooth gradients of quadratic loss for small errors and robustness of linear loss for large errors, effectively combining the advantages of MSE and L1-norm losses. Additionally, to enhance the model's understanding of human body structure, we introduce an additional bone length constraint loss $\mathcal{L}_B$ to ensure that predicted keypoints satisfy the physical constraints of human bones:

\begin{equation}
\mathcal{L}_B = \frac{1}{N}\sum_{i=1}^{N}\frac{1}{|\mathcal{E}|}\sum_{(i,j) \subset E} h(\|\hat{p}_i - \hat{p}_j\|_2 - \|p_i - p_j\|_2),
\end{equation}
where $\mathcal{E}$ is the predefined set of 14 bone connections. The bone loss uses a smaller smoothing parameter to increase sensitivity to bone length changes. The final loss function is:

\begin{equation}
\mathcal{L} = \mathcal{L}_H + \lambda \mathcal{L}_B,
\end{equation}
where $\lambda$ is the bone constraint weight. This design ensures keypoint position accuracy while considering the structural rationality of the human skeleton, achieving excellent results in experiments.

\section{Dataset Description}

\begin{figure}[t]
\centerline{\includegraphics[width=\columnwidth]{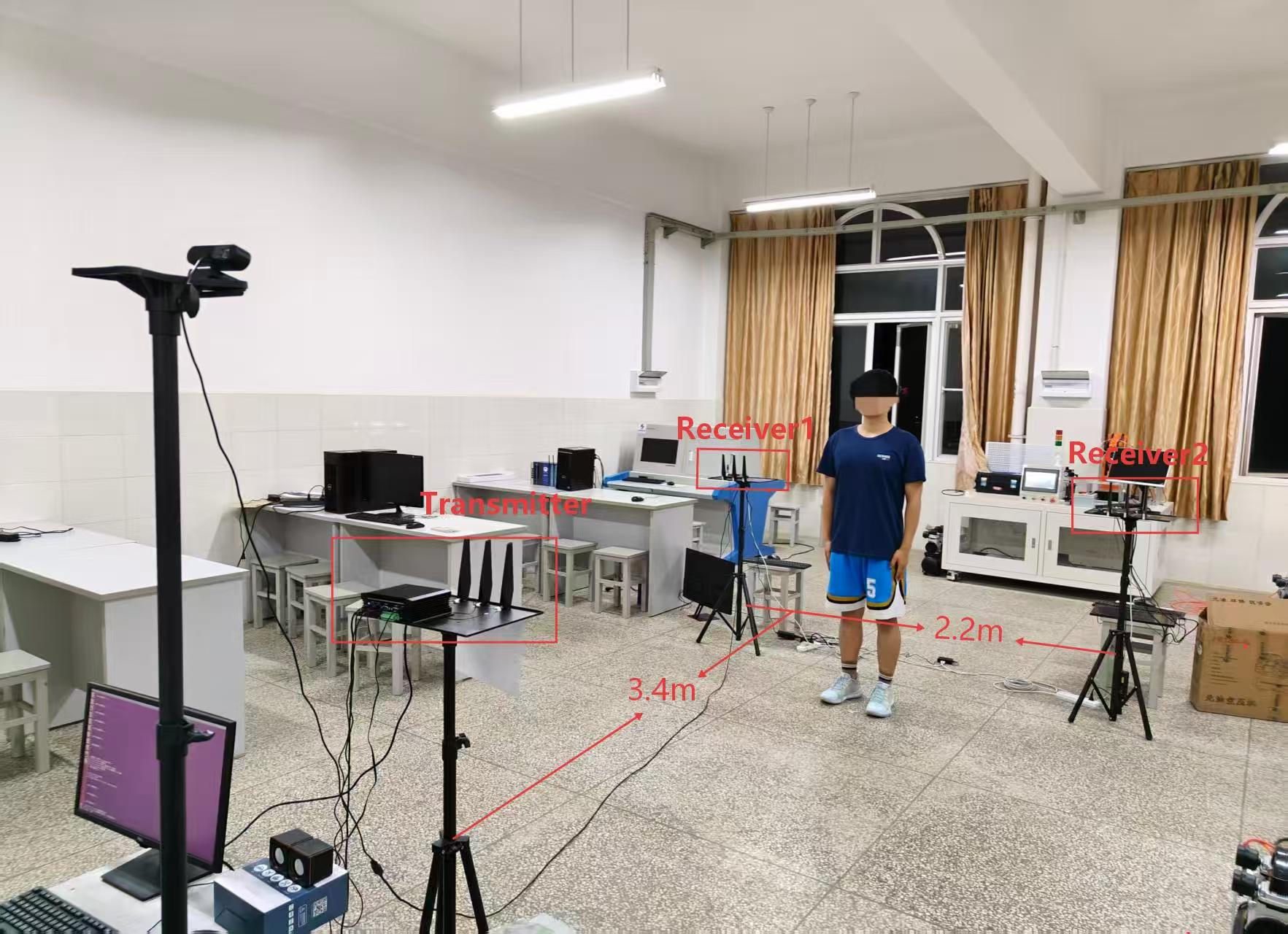}}
\caption{Experimental environment layout demonstration.}
\label{fig:experiment_layout}
\end{figure}

To construct a high-quality WiFi sensing dataset, data collection was conducted in an indoor environment with equipment deployed in a specific triangular topology, as illustrated in Fig.~\ref{fig:experiment_layout}. Specifically, the system consists of one transmitter positioned at the vertex and two receivers at the base. The distance between the two receivers is set to $2.2\,\text{m}$, while the perpendicular distance from the transmitter to the baseline connecting the two receivers is $3.4\,\text{m}$.

Each node is equipped with an Intel 5300 Network Interface Card (NIC) and uses three omnidirectional antennas for signal transmission and reception. All three devices utilize the Linux 802.11n CSI Tool~\cite{halperin2011tool} to extract CSI. The transceivers operate in the 5 GHz frequency band with a 20 MHz channel bandwidth. To ensure temporal alignment across multi-source data, the two receivers are synchronized using the Network Time Protocol (NTP). Additionally, a monocular camera is connected to one of the receiver nodes, allowing us to strictly synchronize the CSI streams and video frames based on recorded timestamps.

The dataset involves 5 volunteers with diverse physical characteristics (height, weight) and clothing styles. During acquisition, subjects performed continuous sequences of 8 daily activities: Walking, Raising Hands, Squatting, Hands Up, Kicking, Waving, Turning, and Jumping, with each action instance lasting approximately 3 seconds. The CSI sampling rate is set to 600 Hz, while the video is recorded at 30 FPS. Consequently, one video frame aligns with a window of 20 CSI packets. The final dataset comprises a total of 360,000 synchronized video-CSI sample pairs. Ground truth generation relies on OpenPose, which extracts 15 2D skeletal keypoints from each video frame, including the nose, neck, shoulders, elbows, wrists, hips (left, right, center), knees, and ankles. These coordinates serve as supervision labels for network training and evaluation.

However, vision-based annotations have inherent limitations, particularly during scenarios involving self-occlusion where OpenPose may fail to detect keypoints, resulting in "zero-coordinate" artifacts $\mathbf{p}_t = \mathbf{0}$. Training directly with these noisy labels could introduce algorithmic biases into the WiFi model. To mitigate this, we implemented a temporal consistency cleaning mechanism during the data preprocessing phase. Specifically, for any missing keypoint at frame $t$, the algorithm locates the nearest valid coordinates $\mathbf{p}_{t_{prev}}$ at frame $t_{prev}$ and $\mathbf{p}_{t_{next}}$ at frame $t_{next}$. The missing coordinate $\hat{\mathbf{p}}_t$ is then reconstructed via linear interpolation:
\begin{equation}
\hat{\mathbf{p}}_t = (1 - \alpha) \cdot \mathbf{p}_{t_{prev}} + \alpha \cdot \mathbf{p}_{t_{next}}
\end{equation}
where the temporal interpolation factor $\alpha$ is defined as:
\begin{equation}
\alpha = \frac{t - t_{prev}}{t_{next} - t_{prev}}
\end{equation}
This strategy not only repairs label errors caused by visual occlusions but also ensures the kinematic smoothness of the pose sequence, thereby providing WiFlow with supervision signals that are more robust and continuous than the raw OpenPose outputs.

\begin{figure*}[t]
\centering
\includegraphics[width=0.95\textwidth]{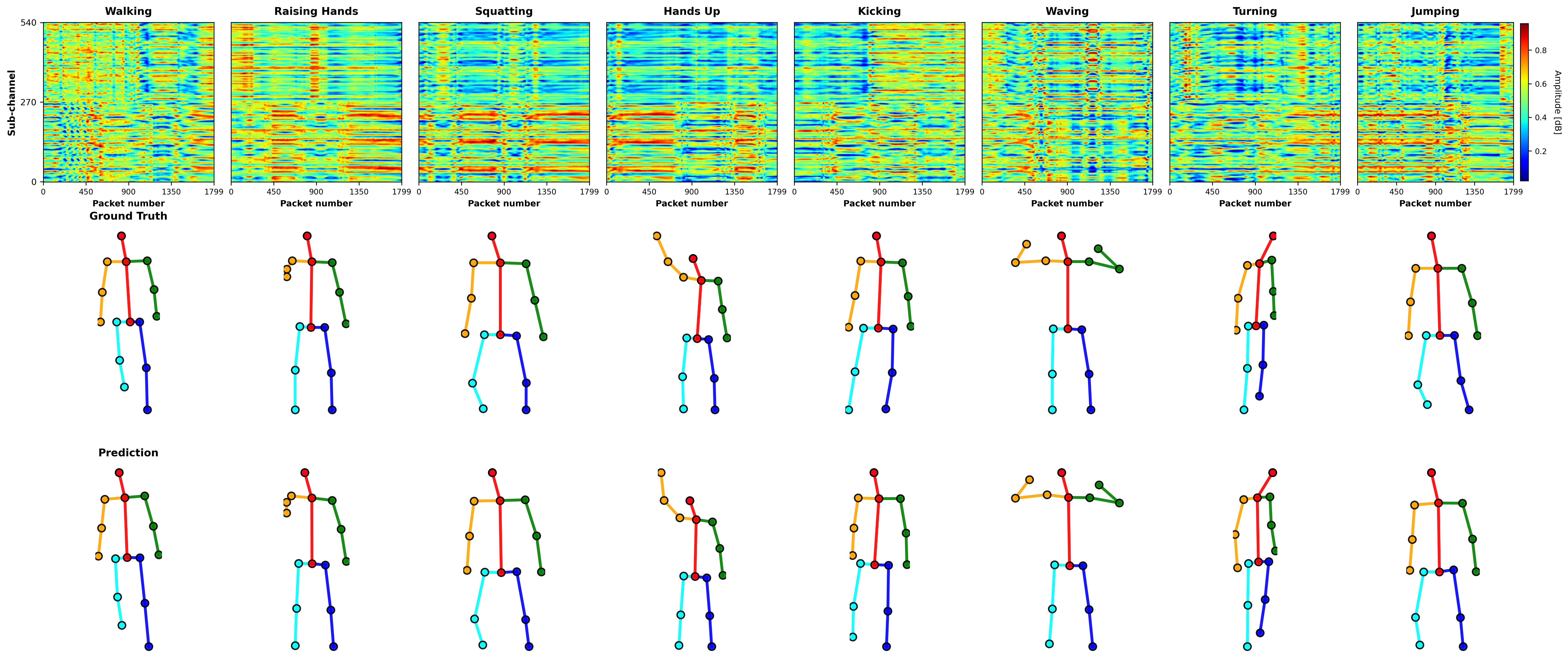}
\caption{Visual comparison of WiFi-based human pose estimation for eight daily actions. 
(Top row) Raw WiFi CSI amplitude heatmaps showing the temporal evolution of signal patterns across 540 sub-channels during each complete 3-second action sequence. 
(Middle row) Ground truth skeletal poses from vision-based model (OpenPose) captured at the middle frame of each action. 
(Bottom row) Corresponding predicted poses from our WiFlow model. 
The eight actions from left to right are: Walking, Raising Hands, Squatting, Hands Up, Kicking, Waving, Turning, and Jumping.}
\label{fig:8actions_comparison}
\end{figure*}

To rigorously assess the performance of WiFlow and address concerns regarding generalization, we introduce two distinct data splitting settings, inspired by established benchmarks such as MM-Fi~\cite{yang2023mm}:

\textbf{Setting 1: Random Split (Subject-Dependent).} 
This setting evaluates the model's capability to learn precise pose representations within a known test subject group. We pool data from all subjects and randomly shuffle it at the \textit{session level} (i.e., by complete file/action sequence). The dataset is then partitioned into training, validation, and testing sets with a ratio of 70\%:15\%:15\%. Crucially, this session-level splitting ensures that frames from the same continuous action sequence do not appear across different sets, thereby preventing temporal data leakage.

\textbf{Setting 2: Cross-Subject Split (Subject-Independent).} 
To evaluate the model's generalization ability to unseen subjects, we employ a \textit{Leave-One-Subject-Out (LOSO)} cross-validation strategy. Specifically, we conduct a 5-fold cross-validation. In each fold, all data from one designated subject is reserved exclusively for testing. The data from the remaining four subjects constitutes the training pool. To maximize data utilization while monitoring overfitting, this training pool is further randomly split at the file level into 90\% for training and 10\% for validation. The final performance metrics are reported as the average across all five folds.

\section{Results and Discussion}

WiFlow is implemented in PyTorch and trained using the AdamW optimizer with weight decay of $5 \times 10^{-5}$. The dataset is split 70\%/15\%/15\% for training, validation, and testing respectively. We train for 50 epochs with batch size 64 and initial learning rate $1 \times 10^{-4}$. A ReduceLROnPlateau scheduler reduces the learning rate by 0.5 when validation MPJPE plateaus for 3 epochs, with minimum learning rate $1 \times 10^{-7}$. The loss function uses Smooth L1-norm with $\beta = 0.1$ and bone constraint weight $\lambda_{\text{bone}} = 0.2$. 
Detailed architecture parameters are shown in Table~\ref{tab:model_params}.

\subsection{Visualization and Analysis}

We compared the visualization results of ground truth labels output by OpenPose (vision-based solution) and predicted labels from WiFlow based on WiFi CSI. Fig.~\ref{fig:8actions_comparison} shows the comparison between ground truth labels from OpenPose (middle) and poses predicted by WiFlow (bottom). It can be seen that WiFlow can achieve performance comparable to vision-based methods. In regular actions such as walking and raising hands, the model has high prediction accuracy for the torso and lower limbs, with keypoint positions basically matching the ground truth labels and the skeleton structure remaining natural. Although WiFlow still has certain limitations in limb prediction, including arm angle deviations and insufficient leg bending, the poses predicted by WiFlow visually maintain reasonable human body structure constraints without obvious joint dislocation or unnatural distortion, validating the effectiveness of the bone length constraint loss in maintaining pose rationality.

\subsection{Comparative Experiments}
WiFlow was evaluated against 4 state-of-the-art baselines: Transformer-based WPformer~\cite{zhou2023metafi++}, ResNet-based WiSPPN~\cite{wang2019can}, U-Net-Transformer-based PerUnet~\cite{zhou2022perunet} and HPE-Li~\cite{HPELI2025}, a lightweight CNN model emphasizing small model size. We first report results on our own dataset under two protocols: User-Dependent (Random Split) and User-Independent (Cross-Subject), followed by a comparison on the public MM-Fi dataset~\cite{yang2023mm}.

\begin{table}[t]
\caption{Model Parameters}
\label{tab:model_params}
\centering
\begin{tabular}{l|c c}
\hline
\textbf{Block} & \textbf{Output size} & \textbf{Parameters} \\
\hline
Input & $540 \times 20$ & - \\
\hline
TCN Layer 1 & $540 \times 20$ & $[1 \times 540] \times 2$, dilation=1 \\
\hline
TCN Layer 2 & $440 \times 20$ & $[1 \times 440] \times 2$, dilation=2 \\
\hline
TCN Layer 3 & $340 \times 20$ & $[1 \times 340] \times 2$, dilation=4 \\
\hline
TCN Layer 4 & $240 \times 20$ & $[1 \times 240] \times 2$, dilation=8 \\
\hline
ConvBlock1 (up) & $8 \times 20 \times 240$ & $[1 \times 3, 8] \times 3$ \\
\hline
ResBlock 1 & $8 \times 20 \times 120$ & $[1 \times 3, 8] \times 3$, stride=(1,2) \\
\hline
ResBlock 2 & $16 \times 20 \times 60$ & $[1 \times 3, 16] \times 3$, stride=(1,2) \\
\hline
ResBlock 3 & $32 \times 20 \times 30$ & $[1 \times 3, 32] \times 3$, stride=(1,2) \\
\hline
ResBlock 4 & $64 \times 20 \times 15$ & $[1 \times 3, 64] \times 3$, stride=(1,2) \\
\hline
AxialAttention & $64 \times 15 \times 20$ & Groups=8, Layer=1 \\
\hline
Decoder & $2 \times 15 \times 20$ & $[3 \times 3, 32], [1 \times 1, 2]$ \\
\hline
Avg Pooling & $2 \times 15 \times 1$ & - \\
\hline
Output & $15 \times 2$ & - \\
\hline
\end{tabular}
\end{table}

\begin{table*}[t]
\caption{Performance Comparison under Setting 1 (Random Split).}
\label{tab:setting1_random}
\centering
\renewcommand{\arraystretch}{1.2}
\begin{tabular}{l|c c c c c c c c}
\hline
Method & PCK@20 & PCK@30 & PCK@40 & PCK@50 & MPJPE(m) & Param (M) & FLOPs (B) & Training Time (h) \\
\hline
WiFlow (Ours) & \textbf{97.25} & \textbf{98.63} & \textbf{99.16} & \textbf{99.48} & \textbf{0.007} & {2.23} & \textbf{0.07} & \textbf{2.30} \\
\hline
WPformer & 70.02 & 82.98 & 89.33 & 93.22 & 0.028 & 10.04 & 35.00 & 35.47 \\
\hline
WiSPPN & 85.87 & 92.23 & 95.52 & 97.48 & 0.016 & 121.50 & 338.45 & 68.10 \\
\hline
PerUnet & 86.11 & 92.34 & 95.45 & 97.29 & 0.016 & 309.09 & 45.92 & 25.50 \\
\hline
HPE-Li & 93.79 & 97.36 & 98.68 & 99.26 & 0.011 & \textbf{0.83} & 1.09 & 3.60 \\
\hline
\end{tabular}
\end{table*}

\subsubsection{Performance on Random Split (Setting 1)}
Table~\ref{tab:setting1_random} reports the results where training and testing data share the same user domain. 
In this scenario, WiFlow achieves significant advantages across all metrics. WiFlow reaches an impressive PCK@20 of 97.25\% and PCK@50 of 99.48\%. The average keypoint error MPJPE is only 0.007 m, reduced by 56.25\% and 75.0\% compared to WiSPPN's 0.016 m and WPformer's 0.028 m, respectively. Compared to WiSPPN using PAM regression, WiFlow improves the strict threshold PCK@20 by 11.38 percentage points. Compared to WPformer with direct coordinate regression, WiFlow improves PCK@20 by 27.23 percentage points. Furthermore, when compared to the lightweight baseline HPE-Li (PCK@20 of 93.79\%), WiFlow still leads by 3.46 percentage points while requiring less than 10\% of its FLOPs (0.07 B vs 1.09 B). This indicates that for personalized applications (e.g., smart homes with fixed members), WiFlow can provide near-perfect pose tracking with minimal parameters.

\begin{table*}[t]
\caption{Performance Comparison under Setting 2 (Cross-Subject / LOSO).}
\label{tab:setting2_loso}
\centering
\centering
\renewcommand{\arraystretch}{1.2}
\begin{tabular}{l|c|c c c c|c}
\hline
\multirow{2}{*}{\textbf{Method}} & \multirow{2}{*}{\textbf{Test Subject}} & \multicolumn{4}{c|}{\textbf{Evaluation Metrics}} & \multirow{2}{*}{\textbf{Training Time (h)}} \\
\cline{3-6}
 & & \textbf{PCK@20} & \textbf{PCK@30} & \textbf{PCK@50} & \textbf{MPJPE (m)} & \\
\hline
\multirow{6}{*}{\textbf{WiFlow (Ours)}} 
 & Subject 1 & 82.65 & 91.54 & 96.80 & 0.023 & 2.45 \\
 & Subject 2 & 90.07 & 96.38 & 98.63 & 0.016 & 3.25 \\
 & \textbf{Subject 3 (Hard)} & \textbf{80.82} & \textbf{89.79} & \textbf{95.93}  & \textbf{0.025} & \textbf{3.50} \\
 & Subject 4 & 90.86 & 95.57 & 98.05 & 0.016 & 3.45 \\
 & Subject 5 & 91.91 & 96.78 & 99.04 & 0.017 & 2.18 \\
 \cline{3-7}
 & \textbf{Average (5-Fold)} & \textbf{87.26} & \textbf{94.01} & \textbf{97.69} & \textbf{0.019} & \textbf{3.17} \\
\hline
WiSPPN & Subject 3 & 71.41 & 82.52 & 92.72 & 0.028 & 51.63 \\
\hline
WPformer & Subject 3 & 68.75 & 81.06 & 92.65 & 0.030 & 137.50 \\
\hline
PerUnet & Subject 3 & 7.70 & 16.06 & 35.82 & 0.109 & 33.40 \\
\hline
HPE-Li & Subject 3 & 79.67 & 89.34 & 95.48 & 0.026 & 3.70 \\
\hline
\end{tabular}\vspace{-5mm}
\end{table*}

\subsubsection{Generalization on Cross-Subject Split (Setting 2)}
Evaluating generalization to unseen users is critical for public deployment. As shown in Table~\ref{tab:setting2_loso}, all methods experience a performance drop due to the domain shift caused by diverse body shapes and gait patterns of new users.

For WiFlow, we conducted a rigorous 5-fold Leave-One-Subject-Out cross-validation. The results show consistent performance across different subjects, with an average PCK@20 of 87.26\%. Notably, Subject 3 emerged as the most challenging case (likely due to distinct motion patterns or body shape), where WiFlow's performance dipped to 80.82\%.

However, even in this "worst-case" scenario, WiFlow significantly outperforms the baselines. Due to the prohibitive computational cost of the baseline methods, we evaluated them using a fixed split with Subject 3 as the test set. Under this identical protocol, WiSPPN and WPformer achieved PCK@20 scores of only 71.41\% and 68.75\%, respectively.

We attribute the sub-optimal performance of WPformer primarily to an architecture-data mismatch. While its design was originally optimized for dense frequency-domain features, our system presents a ``high-dimensional but sparse'' structure, consisting of 18 antenna pairs with only 30 subcarriers each. In such a sparse regime, the shared CNNs and Transformers in WPformer struggle to extract sufficient feature patterns for effective global dependency modeling. In contrast, WiFlow overcomes this limitation by explicitly decoupling spatio-temporal features. The fact that WiFlow maintains a significant margin of 9.41\% and 12.07\% even on the challenging Subject 3 validates its ability to learn intrinsic, transferable pose representations rather than merely overfitting to subject-specific gait patterns.

\begin{table*}[t!]
\caption{Performance comparison on the MM-Fi dataset}
\label{tab:mmFi}
\centering
\renewcommand{\arraystretch}{1.2}
\begin{tabular}{l|c c c c c c}
\hline
\textbf{Method} & \textbf{PCK@20} & \textbf{PCK@30} & \textbf{PCK@40} & \textbf{PCK@50} & \textbf{MPJPE (m)} & \textbf{Param (M)} \\
\hline
WiFlow (Ours) & \textbf{66.73} & \textbf{78.35} & \textbf{84.69} & \textbf{88.46} & \textbf{0.120} & \textbf{1.06} \\
\hline
HPE-Li & 57.35 & 71.70 & 80.16 & 85.74 & 0.138 & 2.06 \\
\hline
PerUnet & 38.11 & 56.13 & 68.62 & 77.00 & 0.193 & 303.97 \\
\hline
WiSPPN & 35.72 & 55.53 & 69.15 & 78.21 & 0.191 & 11.50 \\
\hline
WPformer & 38.54 & 59.73 & 72.30 & 80.24 & 0.183 & 26.52 \\
\hline
\end{tabular}\vspace{-2mm}
\end{table*}

\subsubsection{Computational Efficiency Analysis}
In real-world IoT applications, training efficiency is as crucial as inference accuracy. The last column of Table~\ref{tab:setting2_loso} highlights the immense efficiency advantage of WiFlow.
The Transformer-based WPformer required approximately 137.5 hours (nearly 6 days) to converge on a single fold, while the deep ResNet-based WiSPPN took 51.63 hours. In stark contrast, WiFlow completed the training in only 3.17 hours on average per fold. 
This represents a remarkable 43.4$\times$ speedup over WPformer and a 16.3$\times$ speedup over WiSPPN. Combined with its ultra-low parameter count (2.23 M vs. 121.5 M for WiSPPN), WiFlow establishes a new benchmark for lightweight and efficient WiFi-based human pose sensing.

\subsubsection{Cross-Dataset Comparative Experiments}
To further validate the generalization ability and versatility of WiFlow across heterogeneous sensing environments and different hardware configurations, we conducted supplementary evaluations on the large-scale multimodal wireless sensing dataset MM-Fi~\cite{yang2023mm}. The MM-Fi dataset encompasses 27 complex actions performed by multiple subjects and introduces three evaluation protocols (Protocol 1 with 14 daily actions, Protocol 2 with 13 rehabilitation actions, and Protocol 3 with all 27 actions) alongside two splitting settings (Setting 1 for random split and Setting 2 for cross-subject split). 

For our evaluation, we selected the most challenging combination: Protocol 3 with Setting 1 (P3-S1). Protocol 3 presents highly complex motion patterns across 27 distinct action types. Following our previous cross-subject generalization tests, the adoption of Setting 1 here shifts the evaluation focus toward the model's capacity to generalize across diverse and complex action categories. Consistent with our prior methodology, the random split strictly employs "complete independent action sequences" as the minimum splitting unit to prevent temporal data leakage. Furthermore, since the MM-Fi dataset lacks confidence labels for visual keypoints, Pose Adjacent Matrix (PAM) based on confidence scores cannot be constructed. To ensure absolute fairness, all baseline models in this evaluation were constrained to direct 3D coordinate tensor regression. While this direct regression setup without prior physical topology constraints slightly suppresses the absolute accuracy of PAM-dependent models, it rigorously tests the fundamental representation capability of each architecture to mine spatio-temporal features purely from CSI signals.

As shown in Table~\ref{tab:mmFi}, WiFlow maintains superior performance in this complex 27-action scenario. Under the strict PCK@20 threshold, WiFlow achieves an accuracy of 66.73\% with an MPJPE of 0.120 m. In contrast, WPformer, WiSPPN, and the massive PerUnet (with 303.97 M parameters) struggle on the MM-Fi dataset, fluctuating between 35\% and 38\% for PCK@20 and exhibiting MPJPEs approaching or exceeding 0.190 m. This indicates that model architectures lacking explicit spatio-temporal decoupling mechanisms are highly susceptible to feature confusion when dealing with diverse and dynamically intertwined spatio-temporal motion data. Even when compared to the lightweight baseline HPE-Li, which performs relatively well (57.35\% on PCK@20), WiFlow still maintains a significant lead of +9.38\% on PCK@20 and further reduces the MPJPE by 0.018 m. This rigorous evaluation demonstrates WiFlow's robust generalization and strong feature discrimination capabilities when confronting entirely new datasets with complex action categories.

Regarding model size and computational overhead, the data in Table~\ref{tab:mmFi} further highlights WiFlow's lightweight advantages. After adapting to the input dimensions of the MM-Fi dataset, WiFlow requires only 1.06 M parameters and a mere 0.02 B FLOPs. By comparison, PerUnet's massive 303.97 M parameters do not translate to accuracy gains here, instead incurring a massive computational cost of 42.87 B FLOPs. Similarly, WPformer (26.52 M) and WiSPPN (11.50 M) require 15.87 B and 53.14 B FLOPs, respectively. Notably, the similarly lightweight HPE-Li requires approximately twice the parameters of WiFlow (2.06 M), and lacking an efficient spatio-temporal dimensionality reduction module, its FLOPs (0.35 B) are 17.5 times higher than those of WiFlow.

\subsection{Ablation Experiments}

To verify the effectiveness of each module in the proposed model, we conducted ablation experiments with results shown in Table~\ref{tab:ablation}.

\begin{table}[t!]
\caption{Ablation Study Results}
\label{tab:ablation}
\centering
\begin{tabular}{l|cc}
\hline
Model & PCK@10 & PCK@20 \\
\hline
The proposed model & \textbf{91.36} & \textbf{97.25} \\
\hline
Replace TCN with regular 1D convolution & 84.20 & 96.44 \\
\hline
Replace TCN and Asymmetric Convolution & 83.55 & 95.69 \\
with 2D residual convolution & & \\
\hline
Replace group convolution with & 87.31 & 96.84 \\
depthwise convolution & & \\
\hline
Remove Axial Attention & 91.09 & 97.07 \\
\hline
\end{tabular}
\end{table}

Replacing TCN with regular 1D convolution leads to 7.16\% and 0.81\% drops in PCK@10 and PCK@20. This significant difference indicates that TCN's causal dilated convolution is crucial for temporal modeling. Although regular 1D convolution can also extract temporal features, its limited receptive field cannot capture long-range dependencies within the window. 

Replacing TCN and asymmetric convolution with 2D residual convolution results in 7.81\% and 1.56\% drops, proving the effectiveness of the proposed model's spatio-temporal decoupling architecture. 2D convolution treats CSI frames as images for processing, ignoring the essential differences between temporal and subcarrier dimensions—the temporal dimension has strict causal relationships and continuity constraints, while the subcarrier dimension characterizes the spatial distribution of frequency response. Asymmetric convolution focuses on neighborhood relationships between subcarriers through $1 \times k$ convolution kernels, avoiding information confusion in the temporal dimension.

Replacing group convolution with depthwise convolution leads to a 4.05\% drop in PCK@10, indicating that our decoupled architecture achieves a better balance between parameter efficiency and effective cross-group subcarrier screening.

Finally, removing axial attention causes a slight drop of 0.27\% and 0.18\% in PCK@10 and PCK@20, respectively. After spatio-temporal feature encoding, the features of keypoints already have clear semantic meanings. Axial attention maintains the ability to model global structure while significantly reducing computational complexity through decomposition strategies, playing an important role in maintaining the structural integrity and temporal coherence of poses.

\section{Conclusion}

This paper proposed WiFlow, a WiFi sensing framework for continuous human pose estimation. By constructing a dataset containing 360,000 synchronized samples, we systematically study the pose estimation capability using WiFi CSI on continuous action sequences. The core innovation of WiFlow lies in the network architecture design with spatio-temporal feature decoupling: without destroying sequence structure, the model effectively captures the temporal features of CSI signals through TCN and achieves precise extraction of spatial features between subcarriers using asymmetric convolution. After encoding keypoint information, it uses axial attention mechanisms to model internal keypoint features and global dependencies between keypoints. Experimental results showed that WiFlow reaches leading levels across multiple metrics. The 97.25\% PCK@20 accuracy and 0.007 m MPJPE demonstrated the model's high-precision recognition capability. Compared to existing approaches, WiFlow not only improves estimation accuracy but also significantly reduces model size and computational cost.

\bibliographystyle{IEEEtran}
\bibliography{bibfile}

@InProceedings{HPELI2025,
author="D. Gian, Toan
and Dac Lai, Tien
and Van Luong, Thien
and Wong, Kok-Seng
and Nguyen, Van-Dinh",
title="{HPE-Li}: {WiFi}-Enabled Lightweight Dual Selective Kernel Convolution for Human Pose Estimation",
booktitle="Computer Vision -- ECCV 2024",
year="2025",
publisher="Springer Nature Switzerland",
address="Cham",
pages="93--111",
}

@article{xu2023robust,
  title={Robust abnormal human-posture recognition using OpenPose and multiview cross-information},
  author={Xu, Mingyang and Guo, Limei and Wu, Hsiao-Chun},
  journal={IEEE Sensors Journal},
  volume={23},
  number={11},
  pages={12370--12379},
  year={2023},
  publisher={IEEE}
}

@inproceedings{kotaru2017position,
  title={Position tracking for virtual reality using commodity WiFi},
  author={Kotaru, Manikanta and Katti, Sachin},
  booktitle={Proceedings of the IEEE Conference on Computer Vision and Pattern Recognition},
  pages={68--78},
  year={2017}
}

@article{cao2019openpose,
  title={Openpose: Realtime multi-person 2d pose estimation using part affinity fields},
  author={Cao, Zhe and Hidalgo, Gines and Simon, Tomas and Wei, Shih-En and Sheikh, Yaser},
  journal={IEEE transactions on pattern analysis and machine intelligence},
  volume={43},
  number={1},
  pages={172--186},
  year={2019},
  publisher={IEEE}
}

@inproceedings{huang2020deepfuse,
  title={DeepFuse: An IMU-aware network for real-time 3D human pose estimation from multi-view image},
  author={Huang, Fuyang and Zeng, Ailing and Liu, Minhao and Lai, Qiuxia and Xu, Qiang},
  booktitle={Proceedings of the IEEE/CVF Winter Conference on Applications of Computer Vision},
  pages={429--438},
  year={2020}
}

@inproceedings{yi2024probsparse,
  title={ProbSparse Attention with Stacked Group Convolution for Wireless Signal-Based Human Activity Recognition},
  author={Yi, Dao and Zhang, Haiwei and Feng, Shaohan and Fang, Jinxiang and Wang, Wenbo},
  booktitle={2024 16th International Conference on Wireless Communications and Signal Processing (WCSP)},
  pages={1349--1354},
  year={2024},
  organization={IEEE}
}

@article{luo2024vision,
  title={Vision transformers for human activity recognition using WiFi channel state information},
  author={Luo, Fei and Khan, Salabat and Jiang, Bin and Wu, Kaishun},
  journal={IEEE Internet of Things Journal},
  volume={11},
  number={17},
  pages={28111--28122},
  year={2024},
  publisher={IEEE}
}

@article{fan2024contactless,
  title={A contactless breathing pattern recognition system using deep learning and WiFi signal},
  author={Fan, Dou and Yang, Xiaodong and Zhao, Nan and Guan, Lei and Arslan, Malik Muhammad and Ullah, Muneeb and Imran, Muhammad Ali and Abbasi, Qammer H},
  journal={IEEE Internet of Things Journal},
  volume={11},
  number={13},
  pages={23820--23834},
  year={2024},
  publisher={IEEE}
}

@article{wang2019can,
  title={Can {WiFi} estimate person pose?},
  author={Wang, Fei and Panev, Stanislav and Dai, Ziyi and Han, Jinsong and Huang, Dong},
  journal={arXiv preprint arXiv:1904.00277},
  year={2019}
}

@article{wang2021point,
  title={From point to space: {3D} moving human pose estimation using commodity {WiFi}},
  author={Wang, Yiming and Guo, Lingchao and Lu, Zhaoming and Wen, Xiangming and Zhou, Shuang and Meng, Wanyu},
  journal={IEEE Communications Letters},
  volume={25},
  number={7},
  pages={2235--2239},
  year={2021},
  publisher={IEEE}
}

@inproceedings{yang2022metafi,
  title={{MetaFi}: Device-free pose estimation via commodity {WiFi} for metaverse avatar simulation},
  author={Yang, Jianfei and Zhou, Yunjiao and Huang, He and Zou, Han and Xie, Lihua},
  booktitle={2022 IEEE 8th World Forum on Internet of Things (WF-IoT)},
  pages={1--6},
  year={2022},
  organization={IEEE}
}

@article{zhou2022perunet,
  title={{PerUnet}: Deep signal channel attention in unet for WiFi-based human pose estimation},
  author={Zhou, Yue and Zhu, Aichun and Xu, Caojie and Hu, Fangqiang and Li, Yifeng},
  journal={IEEE Sensors Journal},
  volume={22},
  number={20},
  pages={19750--19760},
  year={2022},
  publisher={IEEE}
}

@article{zhou2023metafi++,
  title={{MetaFi++}: {WiFi-enabled} transformer-based human pose estimation for metaverse avatar simulation},
  author={Zhou, Yunjiao and Huang, He and Yuan, Shenghai and Zou, Han and Xie, Lihua and Yang, Jianfei},
  journal={IEEE Internet of Things Journal},
  volume={10},
  number={16},
  pages={14128--14136},
  year={2023},
  publisher={IEEE}
}

@inproceedings{jiang2020towards,
  title={Towards {3D} human pose construction using {WiFi}},
  author={{W. Jiang, et al.}},
  booktitle={Proceedings of the 26th Annual International Conference on Mobile Computing and Networking},
  pages={1--14},
  year={2020}
}

@article{zhou2022csi,
  title={{CSI-former}: Pay more attention to pose estimation with {WiFi}},
  author={Zhou, Yue and Xu, Caojie and Zhao, Lu and Zhu, Aichun and Hu, Fangqiang and Li, Yifeng},
  journal={Entropy},
  volume={25},
  number={1},
  pages={20},
  year={2022},
  publisher={MDPI}
}

@article{bai2018empirical,
  title={An empirical evaluation of generic convolutional and recurrent networks for sequence modeling},
  author={Bai, Shaojie and Kolter, J Zico and Koltun, Vladlen},
  journal={arXiv preprint arXiv:1803.01271},
  year={2018}
}

@inproceedings{wang2020axial,
  title={Axial-deeplab: Stand-alone axial-attention for panoptic segmentation},
  author={{H. Wang, et al.}},
  booktitle={European conference on computer vision},
  pages={108--126},
  year={2020},
  organization={Springer}
}

@article{halperin2011tool,
  title={Tool release: Gathering 802.11 n traces with channel state information},
  author={Halperin, Daniel and Hu, Wenjun and Sheth, Anmol and Wetherall, David},
  journal={ACM SIGCOMM computer communication review},
  volume={41},
  number={1},
  pages={53--53},
  year={2011},
  publisher={ACM New York, NY, USA}
}

@article{yang2023mm,
  title={{Mm-Fi}: Multi-modal non-intrusive {4D} human dataset for versatile wireless sensing},
  author={{J. Yang, et al.}},
  journal={Advances in Neural Information Processing Systems},
  volume={36},
  pages={18756--18768},
  year={2023}
}

@article{miao2025wi,
  title={{Wi-Fi} sensing techniques for human activity recognition: Brief survey, potential challenges, and research directions},
  author={Miao, Fucheng and Huang, Youxiang and Lu, Zhiyi and Ohtsuki, Tomoaki and Gui, Guan and Sari, Hikmet},
  journal={ACM Computing Surveys},
  volume={57},
  number={5},
  pages={1--30},
  year={2025},
  publisher={ACM New York, NY}
}

@article{alzaabi2025design,
  title={Design and Evaluation of Volunteer User Trials of Unobtrusive Vital Signs Monitoring for Older People in Care Using Wi-Fi CSI Sensing},
  author={Alzaabi, Aaesha and Saied, Imran and Arslan, Tughrul},
  journal={IEEE Journal of Translational Engineering in Health and Medicine},
  year={2025},
  publisher={IEEE}
}

@article{yan2025wi,
  title={{Wi-SFDAGR}: {Wifi}-based cross-domain gesture recognition via source-free domain adaptation},
  author={{H. Yan, et al.}},
  journal={IEEE Internet of Things Journal},
  year={2025},
  publisher={IEEE}
}

@article{jeong2024ubigest,
  title={UbiGest: Smartphone-Based Ubiquitous Gesture Recognition With Wi-Fi},
  author={Jeong, Seung-Hyun and Shin, Kyeong Su and Park, Jihun and Jo, Sanghyeok and Suh, Young-Joo},
  journal={IEEE Internet of Things Journal},
  year={2024},
  publisher={IEEE}
}
\end{document}